\documentclass{article}

\usepackage{arxiv}

\usepackage[utf8]{inputenc} % allow utf-8 input
\usepackage[T1]{fontenc}    % use 8-bit T1 fonts
\usepackage{hyperref}       % hyperlinks
\usepackage{url}            % simple URL typesetting
\usepackage{booktabs}       % professional-quality tables
\usepackage{amsfonts}       % blackboard math symbols
\usepackage{nicefrac}       % compact symbols for 1/2, etc.
\usepackage{microtype}      % microtypography
\usepackage{lipsum}
\usepackage{graphicx}
\graphicspath{ {./images/} }
\usepackage{algorithm}
\usepackage{algorithmic}
\usepackage{amsmath}
\usepackage{amssymb}
\usepackage{epsfig}
\usepackage{booktabs}
\usepackage{multirow}
\usepackage{makecell}
\usepackage{array}

\title{YourSkatingCoach: A Figure Skating Video Benchmark \\for Fine-Grained Element Analysis}

\author{
 Wei-Yi Chen\thanks{First Author} \\
  Institute of Information Science\\
  Academia Sinica\\
  Taipei, Taiwan \\
  \texttt{wyc0223@iis.sinica.edu.tw} \\
  \And
 Yi-Ling Lin \\
  Institute of Information Science\\
  Academia Sinica\\
  Taipei, Taiwan \\
  \texttt{nling827@gmail.com } \\
   \And
 Yu-An Su \\
  Institute of Information Science\\
  Academia Sinica\\
  Taipei, Taiwan \\
  \texttt{yuansu@iis.sinica.edu.tw} \\
  \And
 Wei-Hsin Yeh \\
  Institute of Information Science\\
  Academia Sinica\\
  Taipei, Taiwan \\
  \texttt{weihsinyeh168@gmail.com} \\
  \And
  Lun-Wei Ku \\
  Institute of Information Science \\
  Academia Sinica\\
  Taipei, Taiwan \\
  \texttt{lwku@iis.sinica.edu.tw} \\
}

\begin{document}
\maketitle
\begin{abstract}
Combining sports and machine learning involves leveraging machine
learning algorithms and techniques to extract insight from sports-related data such as player statistics, game footage, and other relevant information. Figure skating is one sport that includes fundamental and challenging elements: rapid movement, fast-changing backgrounds, in-air movement, rotation, and artistic expression, which make this sport a good base from which to start and adapt to other sports. However, datasets related to figure skating in the literature focus primarily on element classification and are currently unavailable or exhibit only limited access, which greatly raise the entry barrier to developing visual sports technology for it. Moreover, when using such data to help athletes improve their skills, we find they are very coarse-grained: they work for learning what an element is, but they are poorly suited to learning whether the element is good or bad. Here we propose air time detection, a novel motion analysis task, the goal of which is to accurately detect the duration of the  air time of a jump. We present YourSkatingCoach, a large, novel figure skating dataset which contains 454~videos of jump 
elements, the detected skater skeletons in each video, along with the gold labels of the start and ending frames of each jump, together as a video benchmark for figure skating. In addition, although this type of task is often viewed as classification, we cast it as a sequential labeling problem and propose a Transformer-based model to calculate the duration. Experimental results show that the proposed model yields a favorable results for a strong baseline. To further verify the
generalizability of the fine-grained labels, we apply the same process to other sports as cross-sports tasks but for coarse-grained task action classification. Here we fine-tune the classification to demonstrate that figure skating, as it contains the essential body movements, constitutes a strong foundation for adaptation to other sports.
\end{abstract}

% keywords can be removed
%\keywords{First keyword \and Second keyword \and More}

\section{Introduction}
As the collection of sports data continues to expand, sports analysts have begun to
leverage machine learning toward understanding and advising coaches and
players. However, more work is done on element 
segmentation~\cite{DBLP:journals/corr/abs-1903-01945} than on element 
analysis~\cite{liu2020fsd10}. That is, datasets and models serve more to identify what or
where the element is rather than how good the element is, which does little to 
help coaches and athletes. In addition, some properties of sports make element
analysis challenging. For example, when using existing models based on static 
movement, it is difficult to analyze sports where athletes move quickly and 
the camera and the background thus change rapidly~\cite{Dwibedi_2019_CVPR}. 
Figure skating, a sport with many challenging features, is a good subject for
a data benchmark. In figure skating, one of the most important elements
to obtain a high score is jumping, which involves speed, rotation, height,
gesture, and a moving background. Recognized jumps include the Toe Loop,
the Salchow, the Loop, the Flip, the Lutz, and the Axel. Each kind
of jump can further be single, double, triple, or quad, depending on the number of
in-air revolutions. Skaters must generate sufficient vertical velocity---height
or air time---to complete the rotations successfully. As a
result, sufficient air time is one key reference to successful
jumps, and analyzing videos can provide this information to both coaches and
athletes for a concrete direction of improvement. This motivates us to propose
air time detection as a new task and develop AI technologies for it. 

In the literature, there are two major datasets for figure skating:
FSD-10 (Figure Skating Dataset)~\cite{liu2020fsd10} and MCFS 
(Motion-Centered Figure Skating dataset including 271 
videos)~\cite{Liu_Zhang_Li_Zhou_Xu_Dong_Zhang_2021}. However, these are not 
suitable for skill improvement. First, their labels
are coarse-grained. FSD-10 is for action recognition, and MCFS is for
temporal action segmentation. That is, they are used merely to find skating
elements from video. Though it may be possible to segment out the
jump from MCFS, its purpose remains to extract specific elements from the video; its
labels are not precise enough to indicate in which frame the ice skate leaves the
ice or lands on the ice. Second, both datasets are composed of world championship
competition videos from YouTube instead of training videos of early or
middle career athletes. The scenario is far from general training and
the elements are too high-level. There are no entry-level elements from
training and it is difficult to obtain minimum air times for each element by
analyzing these videos. Third, FSD-10 is no longer available and MCFS includes only skeleton
videos, not the original videos, perhaps due to copyright concerns. 
The lack of the original videos further complicates element analysis,
and some of the automatically generated skeletons in MCFS are problematic. 

As FSD-10 is no longer available, we randomly sampled 50~videos from
MCFS and checked its labels and skeletons to determine whether it could be used
for element analysis. The average difference between an MCFS jump element start
frame and the actual take-off frame is 56.96~frames (at 30~fps, almost 2~seconds),
and the average difference between the end frame and the actual landing frame is 
83.6~frames (more than 2~seconds). This difference is almost as long as the air
time itself. Fourteen of~50 skeletons from OpenPose~\cite{cao2019openpose} are
troublesome, especially considering they frequently are missing the elbow and wrist points, 
which greatly hinders element analysis. In sum, elements gleaned from such data and
models trained on them provide little useful information for
coaches to use in training athletes, and hence still fall far short of the sport technology
they need. 

Thus we present the new benchmark YourSkatingCoach, which
contains 454~videos originally collected by jump type, i.e., with the gold
label of the jump type. In an attempt to develop sports technology that 
provides vital referential element analysis information to coaches and athletes,
we propose a new task---air time detection---for which we provide gold
labels for each jump's take-off and landing frames. The
take-off frame and landing frame are defined based on the
definition of experts from iCoachskating~\cite{airtime}. To streamline preprocessing, 
YourSkatingCoach also provides the skeleton detected for each frame.
We view this task as a temporal, sequential labeling problem and propose a
Transformer-based model for it as a baseline. 

We also evaluate whether the YourSkatingCoach benchmark can serve as the foundation
for the development of sports technology across other sports. Therefore, we select two
datasets for experiments: FineGym~\cite{shao2020finegym} and our boxing dataset.
The sports involved in these datasets also involve high speeds, complex
elements, and various combinations thereof. We employ the same pipeline for both
datasets for the action classification task for these cross-sport experiments.

\begin{figure}[t]
\centering
\includegraphics[width=0.9\columnwidth]{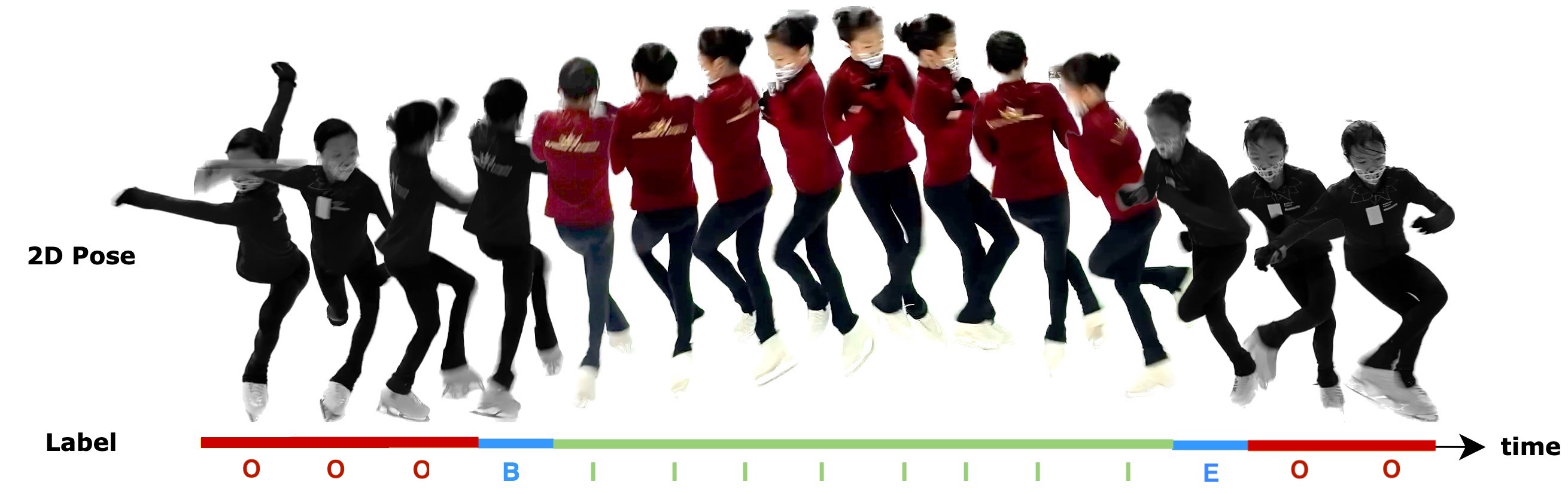} % Reduce the figure size so that it is slightly narrower than the column. Don't use precise values for figure width.This setup will avoid overfull boxes.
\caption{Air time detection of figure skating jump}
\label{fig:air_detection}
\end{figure}

\section{Related Work}
\paragraph{Skeleton-based Representation Learning.}
\cite{DBLP:journals/corr/abs-1912-01001} introduce an approach to
learning probabilistic view-invariant embeddings from 2D skeletons. The learned
embeddings have proven effective on downstream tasks such as
action recognition and video alignment. \cite{DBLP:journals/corr/abs-1801-07455} 
construct an undirected spatial
temporal graph based on a skeleton sequence featuring both intra-body and
inter-frame connections. Their model outperforms the previous state-of-the-art
skeleton-based model on two large-scale datasets. Both of these two approaches
capture motion information in dynamic skeleton sequences without RGB
features. 

\paragraph{Human Professional Sports Datasets.}
Action datasets can be classified into human
action datasets (HADs) and human professional sports datasets (HPSDs). HPSDs play
a crucial role in accelerating the application of machine learning in sports
analytics. An HPSD typically consists of a series of competitive sports actions.
For example, Nevada Olympic sports~\cite{mtlaqa} and MIT Olympic 
sports~\cite{10.1007/978-3-319-10599-4_36} are HPSDs derived from Olympic
competitions. These datasets are proposed for action quality assessment (AQA)
tasks, in which a quality score is provided for a particular action by
analyzing video frames. Most HPSDs are relevant to either
classification or AQA tasks. Here, we focus on the analysis
of figure skating videos for their fast-moving and dynamic characteristics and
propose a novel task that requires models to predict the air time of a
figure skating jump. 

\paragraph{Figure Skating Analysis.}
\cite{liu2020fsd10} propose FSD-10, a fine-grained figure skating dataset,
along with HPS (human pose scatter), a keyframe indicator that
computes the scatter of arms and legs relative to the body. During 
a jump, the HPS value of the in-air frames is claimed to be lower than that of the take-off frames, 
and the extreme point maps a turning point of an action.   % AMH: ?
However, it is difficult to capture the start and end of
the air time given the computed HPS of each frame, since there may be
several extreme points in the plotted jump curve. We address this in our approach 
for air time detection for figure skating by explicitly labeling each
frame as either the start of the flight, inside the flight, the end of the flight, or
other. Another figure skating dataset, the Motion-Centered Figure Skating (MCFS) 
dataset, was introduced by \cite{Liu_Zhang_Li_Zhou_Xu_Dong_Zhang_2021}\ 
MCFS features fine-grained semantics, and is specialized and motion-centered, 
including both RGB-based and
skeleton-based features. However, due to web service restrictions,
FSD-10 is currently not available. Also, we have analyzed the MCFS dataset and find 
that the labels for its take-off and landing frames are not exact enough. 
Hence we constructed YourSkatingCoach, a figure skating dataset containing 454~videos of 
the six recognized jumps.

\section{Methodology}

\begin{figure*}[t]
\begin{center}
    \includegraphics[scale=0.35]{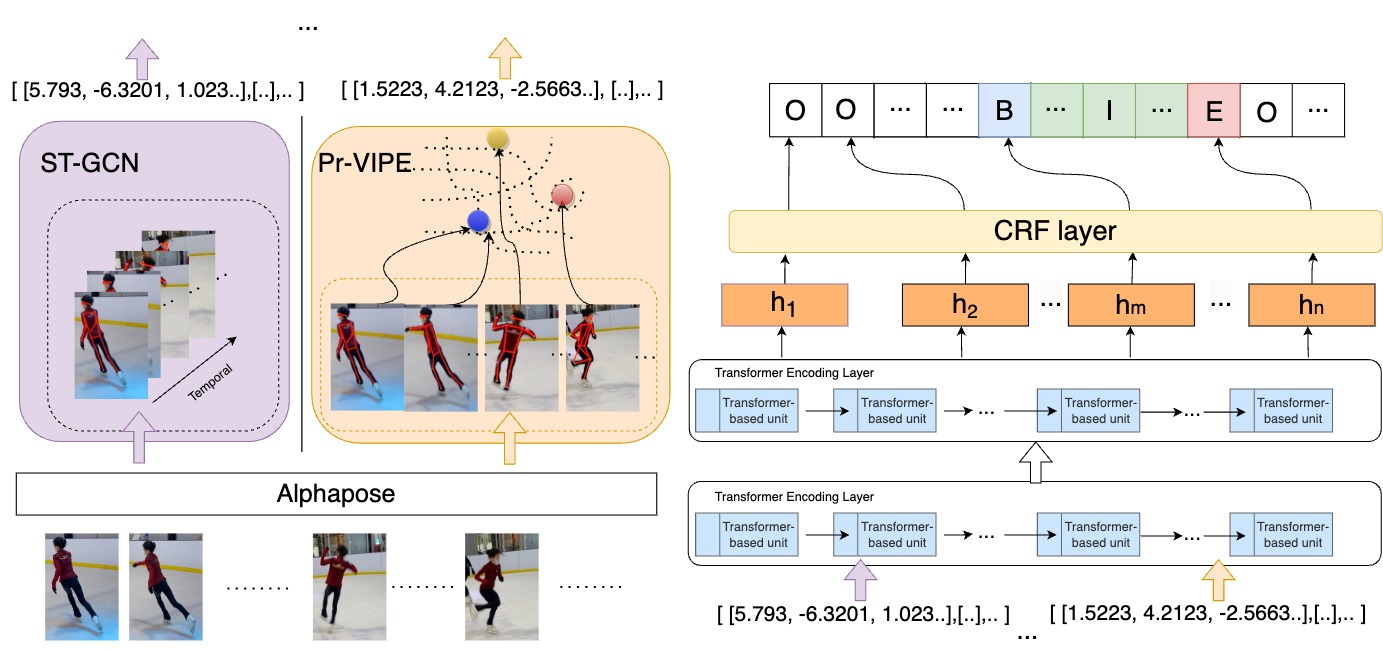}
\end{center}
   \caption{Pipeline of skating and boxing experiments}
\label{fig:skating_pipeline}
\end{figure*}

We cast air time detection as a sequence labeling problem by
classifying each frame as either B\textit{-label}, I\textit{-label}, E\textit{-label}, or O\textit{-label},
which is similar to the BIOES format (Beginning, Inside, Outside,
End, Single) for named entity recognition. A frame is labeled as
B\textit{-label} if it is the take-off frame, I\textit{-label} if it is a
continuous in-air frame, E\textit{-label} if it is the landing frame, or O\textit{-label}
otherwise, as shown in Fig.~\ref{fig:air_detection}. Note that we define
the B frame as precisely that frame in which the skate leaves the ice. The 
air time of the jump is then computed by dividing the number of I\textit{-label}s
by each video's fps (frames per second). The task pipeline is shown in
Fig.~\ref{fig:skating_pipeline}.

\subsection{Model Design}
Here we describe the proposed model in detail. The inputs to the model
are skeleton-based data estimated by the publicly-available 2D pose estimation
algorithm AlphaPose~\cite{alphapose,fang2017rmpe,li2019crowdpose}. 
Let $\Tilde{p_i}$ denote the extracted pose of the
skater for the $i$-th frame. Since the AlphaPose output includes poses of
multiple people, to extract the 2D poses of the skater, we extract the pose with
the highest confidence in the first frame as $\Tilde{p_1}$ while ensuring that
the pose belongs to the skater.  For $P_{i}=\{p_{i,1}, p_{i,2}, ldots\}$, where
$p_{i,j}$ denotes the $j$-th pose for the ${i}$-th frame, we compute the
Euclidean distance between $\Tilde{p_{i-1}}$ and $p_{i,j}$, and obtain the pose
with the minimum distance as $\Tilde{p_i}$. After preprocessing, an input video
sequence is represented as $X=\{x_1,x_2,\ldots,x_T\}$, where $T$ is the number of
frames in the sequence and $x_i = \Tilde{p_i}$ denotes the extracted 2D pose of
the $i$-th frame in the COCO~\cite{cocodataset} format with size (17$\times$2).

\paragraph{Human Pose Embedding.} 
In Pr-VIPE~\cite{DBLP:journals/corr/abs-1912-01001}, a view-invariant embedding
space is learned from 2D joint keypoints. The embeddings were 
applied directly to downstream tasks such as
action recognition and video alignment without further training. In addition
to training with fixed embeddings, we apply the graph CNN model in 
ST-GCN~\cite{DBLP:journals/corr/abs-1801-07455} to form a representation of the
relationships between body parts, where the graph CNN's weights are updated
during training. To extract meaningful features from 2D poses, we use and compare these 
two methods, as discussed in Section~5.3.
These two models take $17\times2$ skeleton sequences as input and project each
skeleton to an $H$-dimension embedding space. The resulting embedding is
then combined with positional embedding and fed to the Encoder-CRF model described
below.

\paragraph{Encoder-CRF for Sequence Labeling.}
The Transformer~\cite{DBLP:journals/corr/VaswaniSPUJGKP17} architecture follows
the encoder--decoder paradigm. The encoder component typically stacks six
identical encoder layers, in which each layer utilizes both a multihead
self-attention mechanism and a position-wise fully connected feedforward
network. With self-attention, the encoder captures
latent semantics and global dependencies from an input sequence of symbol
representations. In order to learn contextual relationships among frames, we
stack two encoder layers in our model. For an input sequence of $T$~frames, the
encoder encodes the context of the frames into their respective $H$-dimension 
representations. The encoded representation of each frame is then linearly
projected onto a layer whose size is equal to the number of distinct labels
$K$, namely, $R^H\rightarrow R^K$. A straightforward approach would be to use the
softmax output from this layer as the predictions. However, there exist strong
dependencies across output labels such that they must
appear in the following order: $O\rightarrow B\rightarrow I\rightarrow
E\rightarrow O$. Therefore, we use a CRF (conditional random 
field)~\cite{laffertyCrf} to utilize neighboring label information when inferring the
final predictions for each frame. The CRF parameters are a state
transition matrix $A\in R^{K\times K}$, where $A_{i,j}$ models the transition
from the $i$-th state to the $j$-th state for a pair of consecutive time steps.
The output scores of the linear layer are then combined with the transition
scores to compute the score of a sequence of label predictions. Let $C$ be the
output scores of the linear layer of size ($T$, $K$), and $C_{i,j}$ denote the
score of the $j$-th label of the $i$-th frame in a video sequence. For a
sequence of predicted labels $y=\{y_1,y_2,\ldots,y_T\}$, we define its score as
described in \cite{DBLP:journals/corr/LampleBSKD16}:
\begin{equation}
    S(X,y)=\sum_{i=1}^{T} A_{y_i,y_{i+1}}+\sum_{i=1}^{T} C_{i,y_i}.
\end{equation}
A probability for sequence $y$ is yielded from a softmax over all possible
sequences:
\begin{equation}
    p(y|X)= \cfrac{e^{S(X,y)}}{\sum_{\Tilde{y}\in Y_X} e^{S(X,\Tilde{y})}}.
\end{equation}
During training, we maximize the log probability of the correct sequence, and at
inference time, we predict the sequence with the maximum score using the Viterbi
algorithm.

\section{Dataset}
Given the unavailability of FSD-10 dataset and the imprecise labeling of the
MCFS start/end labels, we compiled our own figure
skating dataset for use in the experiments here. The dataset contains 408~videos in six
jump categories, where each category consists of
9 to~173 videos. The videos range from 1.3s to~10s, with a frame rate of
30~fps and resolution of 1920$\times$1080. The collected videos are all
practice clips of a 9-year-old female skater, and were categorized by experts.
The start/end frames of the flight phase(s) of each video
were manually annotated by research assistants. There can be 1 to~3 jumps
of the same type in a video. An example of one video is shown in
Table~\ref{tab:video_record}, where $\mathit{start\_flight}_1$ and $\mathit{end\_flight}_1$
represent the start and the end frames of the first jump's flight phase
respectively.  For further experiments, we assembled a separate dataset for the
jump category with more than 40~videos, and the \textit{all\_jump} dataset, which
contains all the videos. In addition to separating the videos by action, we
compiled \textit{single\_jump} and \textit{multiple\_jump} as 
supplementary datasets to examine the influence of the number of jumps on
this task. \textit{Single\_jump} contains videos with only one jump, and 
\textit{multiple\_jump} includes videos with multiple jumps. The dataset details 
are provided in Table~\ref{tab:skating_datasets}.

\begin{table}
\begin{center}{
\begin{tabular} {  c  c  }
\hline
Property & Value \\
\hline
video-ID & 001 \\
category & Axel \\
$\mathit{start\_flight}_1$ & 31 \\
$\mathit{end\_flight}_1$ & 40 \\
\hline
\end{tabular}
}
\end{center}
\caption{A video record}
\label{tab:video_record}
\end{table}

\begin{table*}[ht]
\small
\centering
\addtolength{\tabcolsep}{-3pt}
\resizebox{0.7\linewidth}{!}{%
\begin{tabular} {  c  c  c c c c c c}
\hline
Dataset & Axel & Loop & Flip & Lutz & single\_jump & multiple\_jump & all\_jump \\
\hline
Training videos & 42 & 155 & 108  & 74 & 336 & 72 & 408\\
Videos with $\geq$ 2 jumps & 28 & 10 & 16 & 6 & 0 & 72 & 72\\
Average of frames & 200 & 113 & 119 & 107 & 113 & 189 & 127\\
\hline
Testing videos & 5 & 17 & 12  & 9 & 35 & 11 & 46\\
Videos with $\geq$ 2 jumps & 4 & 2 & 3 & 1 & 0 & 11 & 11\\
Average of frames & 137 & 123 & 114 & 102 & 110 & 153 & 120\\
\hline
Total of videos & 47 & 173 & 120  & 83 & 371 & 83 & 454\\
Videos with $\geq$ 2 jumps & 32 & 12 & 18 & 7 & 0 & 83 & 83\\
Average of frames & 194 & 114 & 118 & 106 & 113 & 185 & 126\\
\hline
\end{tabular}
}
\caption{The details of the figure skating datasets.}
\label{tab:skating_datasets}
\end{table*}

\paragraph{Data Augmentation.}
Given the time-consuming nature of video collection and annotation, we
generated new data points using the proposed method to amplify the dataset 
and avoid overfitting. We observed that take-off and landing actions
take up approximately 30~frames in a 30-fps setting. Therefore, we extracted 
multiple data points from a single video by trimming out the
context from the beginning and end while making sure that at least 
30~frames of context were kept. The number of training samples after data
augmentation is displayed in Table~\ref{tab:augmented_datasets}.

\begin{table}[ht]
\fontsize{9}{10}\selectfont
\centering
\begin{tabular} {c c}
\hline
Dataset & Training samples \\
\hline
Axel & 1544 \\
Loop & 1190 \\
Flip & 914 \\
Lutz & 460 \\
single\_jump & 2745\\
multiple\_jump & 4950 \\
\hline
\end{tabular}
\caption{The number of training samples for each dataset after data augmentation.}

\label{tab:augmented_datasets}
\end{table}

\section{Experiments}
In this section, we 
% evaluate the air time  
  estimate air time duration   % AMH: check (ok!)
using the YourSkatingCoach benchmark.
In particular, we list two baseline results when using the pretrained Pr-VIPE
and trainable graph CNN (both commonly-seen practices) for our human
embedding. In addition, we also conduct cross-action experiments in which we
test the model trained on the samples of a specific jump using samples of other
jumps. 
%In the following subsections, we describe the details of our datasets and implementation, and discuss the results.

\subsection{Evaluation Metrics}
As the task goal is to detect the air time of jumps in a video, we propose using
the mean error percentage to evaluate the quality of the predictions
in addition to the commonly used accuracy, F1 score, and edit
distance metrics. An overlapping prediction is a prediction whose air time
overlaps with that of the ground truth. The mean error percentage is
computed as
\begin{equation}
\cfrac{
\sum^{N}_{i=1}
\cfrac{|\mathrm{len}(\mathit{pred}_i) - \mathrm{len}(\mathrm{ans}(\mathit{pred}_i))|}{\mathrm{len}(\mathrm{ans}(\mathit{pred}_i))}
}
{N},
\end{equation}
where $N$ denotes the number of overlapping predictions, $\mathit{pred}_i$ denotes the
$i$-th prediction, $\mathrm{len}(\mathrm{ans}(\mathit{pred}_i))$ is the ground truth flight time, and
$\mathrm{len}(\mathit{pred}_i)$ is the flight time of $\mathit{pred}_i$. 

\subsection{Implementation Details}

\paragraph{Pr-VIPE.} 
In the experiments, we used the released Pr-VIPE checkpoint, which is trained on
the Human3.6M dataset, with an embedding size of~16. To generate
Pr-VIPE embeddings for 2D poses predicted by AlphaPose, we transformed the joint
keypoints from the COCO format to the Human3.6M format during preprocessing. 

\paragraph{GCN.}
For the skeleton graph $G=(V,E)$ used in this experiment, there were 17 joint
nodes~$V$, along with the skeleton edges~$E$. We constructed the graph CNN model
with spatial configuration partitioning as described in
\cite{DBLP:journals/corr/abs-1801-07455}. The model extracts features from 
17~joint coordinates for each frame.

\paragraph{Training Parameters.}
We trained our models with a batch size of~128, a learning rate of $1e-4$, and
used Adam as the optimizer for 200~epochs.

\subsection{Results}
We evaluated the quality of air time estimation using Pr-VIPE embeddings by the mean
error percentage. As a comparison, we also used trainable GCN as a feature
extractor.  The results are shown in
Table~\ref{tab:flight_detection_comparison}. Pr-VIPE, the pretrained
model, and GCN, directly intergrated and fine-tuned 
in  % AMH: check (ok)
our model, are two major approaches to generating skeleton embeddings for model training.
We show that in this task with the YourSkatingCoach benchmark, their
performance is comparable. However, the large gap in the F1 score shows that
GCN is much better than Pr-VIPE. Further analysis showed that
when using Pr-VIPE generated embeddings, our model has less confidence to
predict air tags and hence predicts fewer. In practice, 
Pr-VIPE can be used if the application needs the model to predict more strictly, and
GCN can be used if the model should predict in a more lenient way.

\begin{table}[ht]
\fontsize{9}{11}\selectfont
\centering
\small  % You can adjust the size to match the surrounding text
\begin{tabular}{p{0.6\linewidth}}
\specialrule{.1em}{.05em}{.05em}
\textbf{Methods:} Pr-VIPE + Encoder-CRF \\
\textbf{Accuracy (\%):} 95.3 \\
\textbf{F1-score:} 0.564 \\
\textbf{Mean Error Percentage (\%):} 27.17 \\
\textbf{Edit Distance:} 5.674 \\
\specialrule{0.1em}{.05em}{.05em}
\textbf{Methods:} GCN + Encoder-CRF \\
\textbf{Accuracy (\%):} \textbf{96.3} \\
\textbf{F1-score:} \textbf{0.671} \\
\textbf{Mean Error Percentage (\%):} \textbf{25.05} \\
\textbf{Edit Distance:} \textbf{4.435} \\
\specialrule{.1em}{.05em}{.05em}
\end{tabular}
\caption{Comparison of air time detection results of the Pr-VIPE+Encoder-CRF and GCN+Encoder-CRF trained and tested on the  all\_jump dataset.}
\label{tab:flight_detection_comparison}
\end{table}

\paragraph{Cross-Action Experiments.}
To explore the air time performance on different jumps, we conducted a
cross-action experiment to train on one jump but test on another.
In Table~\ref{tab:embeddingformer_flight_detection}, the model trained
on the \textit{all\_jump} dataset emerges as the most capable air time detector
since it performs the best in all datasets except Axel, and is still a
close second best for the Axel jump.  Note that 
single-jump-trained models other than the Axel model perform unsatisfactorily on
the Axel jump. We list the 10~videos with the lowest
accuracy for each cross-action model and count the occurrences of each jump in
these videos: Axel is the most confusing jump. This is straightforward, as
Axel is the only forward take-off jump which therefore also has a different number
of revolutions. This also shows that when making predictions, the model considers not only the
take-off spot but other parts of the body as well. This
indicates a direction for model enhancement: to improve performance, we could add an attention
mechanism on the body part in focus in the current task.

Another reason observed for the uneven prediction quality is that 
over $25\%$ videos in the test split contain multiple jumps, i.e.,
combination jumps, as shown in Table~\ref{tab:skating_datasets}. One 
reason why videos containing multiple jumps are challenging is that the
interval between jumps can vary and could be very short, which changes the
context. Multiple jumps can be done as a combination in which each
jump is immediately followed by another jump, or there can be other figure
skating elements in between. 

\begin{table*}[t]\tiny
\resizebox{\textwidth}{!}{%
\begin{tabular}{|c|c|c|c|c|c|c|}
\hline
\multirow{3}{*}{\textbf{Training dataset}} & 
    \multirow{3}{*}{\textbf{Metrics}} & 
    \multicolumn{5}{|c|}{\textbf{Testing dataset}} \\
\cline{3-7}
& & \textbf{Axel (5)} & \textbf{Loop (17)} & \textbf{Flip (12)} & \textbf{Lutz (9)} & \textbf{All Jump (46)} \\
\hline
\multirow{4}{*}{\makecell{Axel \\ (42)}} & accuracy (\%) & \textbf{94.3} & 95.8 & 92.0 & 93.6 & 94.4 \\
    & mean error (\%) & \textbf{26.63} & 51.16 & 55.47 & 62.27 & 49.16 \\
    & macro avg f1 & \textbf{0.627} & 0.513 & 0.470 & 0.453 & 0.526 \\
    & avg edit distance & \textbf{7.800} & 5.176 & 9.167 & 6.556 & 6.761 \\
\hline
\multirow{4}{*}{\makecell{Loop \\ (156)}} & accuracy (\%) & 89.8 & 95.9 & 94.4 & 92.5 & 94.4 \\
    & mean error (\%) & 44.27 & 28.40 & \textbf{33.42} & 46.99 & 34.36 \\
    & macro avg f1 & 0.438 & 0.607 & 0.462 & 0.428 & 0.538 \\
    & avg edit distance & 14.000 & 5.059 & 6.417 & 7.667 & 6.804 \\
\hline
\multirow{4}{*}{\makecell{Flip \\ (108)}} & accuracy (\%) & 87.8 & 95.7 & 94.0 & \textbf{97.1} & 94.9 \\
    & mean error (\%) & 75.86 & 25.56 & 38.00 & \textbf{10.22} & 36.61 \\
    & macro avg f1 & 0.338 & 0.564 & 0.637 & 0.567 & 0.556 \\
    & avg edit distance & 16.800 & 5.294 & 5.250 & \textbf{3.000} & 6.174 \\
\hline
\multirow{4}{*}{\makecell{Lutz \\ (74)}} & accuracy (\%) & 90.6 & 95.4 & 91.7 & 94.2 & 93.8 \\
    & mean error (\%) & 61.91 & 35.24 & 57.66 & 44.53 & 46.82 \\
    & macro avg f1 & 0.415 & 0.517 & 0.430 & 0.471 & 0.483 \\
    & avg edit distance & 13.000 & 5.706 & 9.500 & 6.000 & 7.500 \\
\hline
\multirow{4}{*}{\makecell{All Jump \\ (408)}} & accuracy (\%) & 92.6 & \textbf{97.5} & \textbf{96.1} & 96.4 & \textbf{96.3} \\
    & mean error (\%) & 30.54 & \textbf{15.95} & 33.69 & 24.90 & \textbf{25.05} \\
    & macro avg f1 & 0.555 & \textbf{0.714} & \textbf{0.702} & \textbf{0.676} & \textbf{0.671} \\
    & avg edit distance & 10.200 & \textbf{3.059} & \textbf{4.417} & 3.667 & \textbf{4.435} \\
\hline
\end{tabular}
}
\caption{Flight detection performance of GCN + Encoder-CRF. The table demonstrates the performance of the models trained and tested on different actions. The sizes of the datasets are listed in parentheses.}
\label{tab:embeddingformer_flight_detection}
\end{table*}

\paragraph{Single- and Multiple-Jump Experiments.}
To investigate the impact of the number of jumps in the videos for this task,
we trained the GCN+Encoder-CRF model on the \textit{single\_jump} dataset and tested it on
both the \textit{single\_jump} and \textit{multiple\_jump} datasets. In
Table~\ref{tab:single_multiple_jump_experiments} we see that when all the videos
in the dataset contain only one jump, the model performs well, achieving an
accuracy of 96.7\% and a mean error percentage of 25.46\%. However,
when all the testing videos contain multiple jumps, the model struggles to
predict the air time, resulting in an accuracy of 90.7\% and a mean error
percentage of 61.84\%. These results indicate that the next issue to study is 
how to mitigate the length issue (the number of jumps and the interval lengths).

\begin{table}[ht]
\centering
\begin{tabular}{ c|c | c}
\hline
Testing dataset & {single\_jump} & {multiple\_jump}\\
\hline
Accuracy (\%) & \textbf{96.7} & {90.7} \\
F1-score\textbf{} &\textbf{0.615} & {0.466}\\
Mean Error Percentage (\%)&\textbf{25.46} & {61.84} \\
{Edit Distance} & \textbf{3.629}&{14.364}\\
\hline
\end{tabular}
\caption{Results of the GCN with Encoder-CRF model trained on the single\_jump and tested on the multiple\_jump dataset.}
\label{tab:single_multiple_jump_experiments}
\end{table}

\begin{figure}[ht]
\begin{center}
\resizebox{0.6\linewidth}{!}{%
\includegraphics{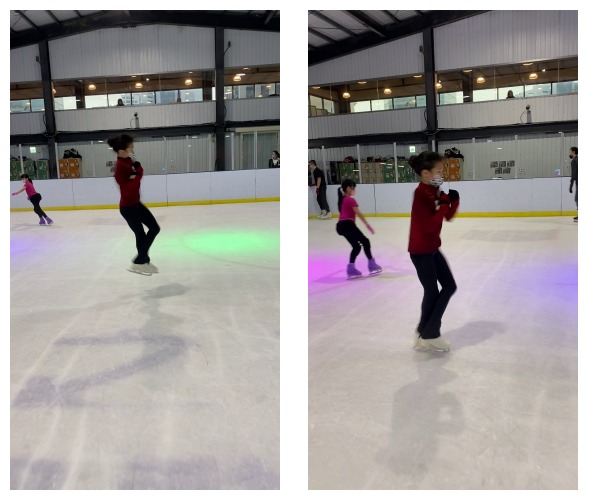}
}
\end{center}
	\caption{Left: A picture taken during a jump. Right: A picture taken during
	a spin.}
\label{fig:jump_spin}
\end{figure}

\paragraph{Other Issues.}
Spins are another important element in figure skating in which skaters rotate
in one spot. Both spins and jumps involve rotations: spins
involve rotations on the ground and jumps in the air, as illustrated in
Fig.~\ref{fig:jump_spin}. We test our cross-action models with a video
containing a jump and a spin to see if the model can distinguish them. The results
show that some models fail to differentiate between these two types of
rotations, in that they predict the spin as the air time of a jump. 

\section{Cross-Sport Experiments}
In this section, we seek to validate the capabilities of the proposed model for
air time detection applied to other sport tasks. We adopt a simple
fine-tuning method for these cross-sport experiments. We use the FineGym dataset and
our boxing dataset. Here we leverage 
action classification for other sports as the downstream task of figure
skating.

\subsection{Action Classification}
For action classification for other sports, we follow the same pipeline used in
the skating experiment illustrated in Fig.~\ref{fig:skating_pipeline}.
Note however that air time detection is a sequence labeling task but 
action classification is a sequence-to-action task. That is, here we input the
player's skeleton sequence into the model, and the model predicts the
corresponding action for the sequence. Therefore, we replace the CRF layer in
our framework with a classification output layer after the Transformer encoder.
For evaluation, we adopt two commonly used metrics: accuracy, F1 score.

\subsection{Dataset}
Our boxing data was collected in a simulated boxing class scenario. In
these videos, the camera team instructed the boxer to stand in the center and
perform ``jabs'' and ``crosses'', both boxing actions. We collected one video from four
cameras from different angles for ten boxers, yielding a total of 40~videos,
after which we utilized AlphaPose to extract the boxer's 2D joints. 
The FineGym dataset contains four different sport events: balance
beam (BB), floor exercise (FX), uneven bars (UB), and vault (VT). Each
event has action categories except for the VT event. For example, BB can be
classified into actions such as BB\_leap\_jump\_hop, BB\_turns,
BB\_flight\_salto, and BB\_flight\_handspring. The FineGym
dataset also provides labels that include action start and end times (but
not the precise frame), action category, and video resources. In the
FineGym dataset, there are initially 303 videos. However, due to some
videos being currently unavailable, we were only able to find 246~videos.

\paragraph{Issues when Using Competition Videos.}
As with \mbox{FSD-10} and MCFS, the videos in the FineGym dataset are from competitions.
There are challenges when such competition videos are adopted for element analysis.
First, videos contain other people such as audience, judges, and coaches in addition to the
athletes, which complicates skeleton extraction. 
Second, as the video angle changes frequently in a real competition,
it is difficult to consistently track the player's skeleton. To address these
challenges and ensure high quality experiments, 
we put manual effort into the experimental FineGym data for cleaning background noise  % AMH: this is a bit too vague -- what manual effort? (ok!)
to make sure the skeletons and labels were consistent
and correct for the target athlete.

With our boxing dataset the story is different: the cost is high for collecting
training-style data. We invited boxers to our facilities for shooting. Due to the
constraints of both boxers and the camera team, as well as the 
equipment problems such as cameras overheating, few videos could be obtained
each time. 

Therefore, we used data preprocessing and augmentation for the boxing dataset.
% \paragraph{Data Preprocessing.}
% Regarding the boxing dataset, as mentioned earlier, we have 40 videos. However, we will record only one video per different boxer, with the sole distinction being the shooting angle. Consequently, data augmentation is necessary. Additionally, 
The dataset was divided randomly into 55\% for training and 45\% for
validation. As each video in FineGym contains multiple sport events, we divided
these videos by sport events. Sport event segmentation resulted in 1260~BB videos,
1260 FX videos, and 854 UB videos. We further
partitioned each video by actions. For example, the UB sport event includes
actions such as circles, flight\_same\_bar, and transition\_flight. The details
of the processed dataset are shown in Table~\ref{tab:cross_sport_data_description}.

\begin{table}[ht]
\centering
\fontsize{9}{12}\selectfont{%
\begin{tabular} { c c c c c}
\hline
Dataset & Total   & Training  & Testing  & Type of Action\\
\hline
FineGym(BB) & 1260 & 880 & 380 & 4 \\
FineGym(UB) & 854 & 596 & 258 & 3 \\
FineGym(FX) & 1260 & 880 & 380 & 4 \\
FineGym(All) & 3374 & 2356 & 1018 & 11 \\
Boxing & 8160 & 4520 & 3640 & 2 \\
\hline
\end{tabular}
}% end of resizebox    
\caption{The details of our cross-sports dataset used for action classification including the number of training samples, the number of testing samples, and the count of action categories.}
\label{tab:cross_sport_data_description}
\end{table}

\begin{table}[t]
\centering 
\fontsize{10}{12}\selectfont{%
\begin{tabular}{c c c c c c}
\hline
Dataset & Method & Accuracy (\%) & F1 score  \\
\hline
\multirow{2}*{FineGym(BB)} & Vanilla & 43.2 & 0.252 \\
& Fine-tuning & \textbf{49.7} & \textbf{0.405} \\
\hline
\multirow{2}*{FineGym(UB)} & Vanilla & 64.1 & 0.459 & \\
& Fine-tuning & \textbf{87.2} & \textbf{0.792} \\
\hline
\multirow{2}*{FineGym(FX)} & Vanilla & 43.2 & 0.252 \\
& Fine-tuning & \textbf{49.7} & \textbf{0.405} \\
\hline
% \multirow{2}*{FineGym(all)} & Vanilla & 41.5 & 0.192 \\
% & Fine-tuning & \textbf{49.0}  & \textbf{0.208} \\
% \hline
\multirow{2}*{Boxing} & Vanilla & 88.1 & 0.876 \\
& Fine-tuning & \textbf{93.1} & \textbf{0.931} &\\
\hline
\end{tabular}
}
\caption{Cross-sport action classification experiment results}
\label{tab:cross_sport_result}
\end{table}

\subsection{Results}
The performance of action classification are superior.
Table~\ref{tab:cross_sport_result} shows that fine-tuning the action classification
for the current sport using the pretrained figure skating model improves the F1 score from 1.5\% to 13.0\%, which demonstrates the power of the fine-grained figure skating data as the foundation for tasks in other sports, with various essential body movements.

We conducted a comprehensive study of the FineGym dataset, augmenting both the training and test data, leading to several noteworthy insights. With an expanded training dataset, performance improves across most sport events. Fine-tuning the model further provides significant enhancements, except for the Balance Beam (BB) event. While performance improves during fine-tuning, BB's performance declines with the incorporation of additional training data. We observed that movements such as handstands and mid-air circling are frequently seen in BB. However, these actions pose challenges for generating accurate skeletons using tools like AlphaPose. Despite manually tracking skeletons to ensure high-quality experiments, issues persist due to the initially poorly generated skeleton in the BB sport.

Both the Balance Beam (BB) and Floor Exercise (FX) are gymnastics events in which gymnasts showcase a variety of skills through dancing and tumbling. Conversely, the "Uneven Bars" (UB) event encompasses three distinct types of movements:
(1) UB\_circles, involving the gymnast's circular action around the uneven bars using both hands.
(2) UB\_flight\_same\_bar, where the gymnast propels themselves from one uneven bar and returns to the original bar.
(3) UB Transition Flight, entailing the transition between two uneven bars.
As illustrated in Table~\ref{tab:cross_sport_result}, the model demonstrates superior performance when tested on UB actions compared to BB and FX events. Given that gymnasts execute UB maneuvers in mid-air, this result further corroborates that the proposed novel task of "air time prediction" significantly enhances the model's performance in downstream tasks involving aerial actions.

\section{Conclusion}
In this paper, we present YourSkatingCoach, a new skating benchmark for figure
skating element analysis. We propose air time prediction as a novel task for use in
improving the major scoring element, jumps, in the
figure skating field. We propose a Transformer-based model as a strong
baseline for this task. Furthermore, we demonstrate that this new task and its
proposed model not only serve as a baseline in the benchmark but even
provide a good foundation for downstream tasks of other sports, given its superior
performance gain in cross-sport experiments, due to the essential body
movement information they provide. These results confirm that the
fine-grained element analysis benchmark is a promising
direction that benefits both athletes and coaches. We will continue to extend
this benchmark in size and variety in the future. Our next goal is to address the challenges of 
rapid movement and background change along with bodies and skeletons to
provide more detail for athletes and coaches.
%-------------------------------------------------------------------------

\end{document}